\documentclass[10pt,twocolumn,letterpaper]{article}

\usepackage{cvpr}
\usepackage{times}
\usepackage{epsfig}
\usepackage{graphicx}
\usepackage{amsmath}
\usepackage{amssymb}
\usepackage{booktabs}       
\usepackage{siunitx}
\usepackage{authblk}

\usepackage{algorithm}
\usepackage[noend]{algpseudocode}

\usepackage[pagebackref=true,breaklinks=true,letterpaper=true,colorlinks,bookmarks=false]{hyperref}

\cvprfinalcopy 

\def\cvprPaperID{7039} 

\ifcvprfinal\fi
\begin{document}

\title{Retina U-Net: Embarrassingly Simple Exploitation\\ of Segmentation Supervision for Medical Object Detection}

\makeatletter
\renewcommand\AB@affilsepx{, \protect\Affilfont}
\makeatother

\author[1] {Paul F. Jaeger}
\author[1] {Simon A. A. Kohl}
\author[2] {Sebastian Bickelhaupt}
\author[1] {Fabian Isensee}
\author[3] {Tristan Anselm Kuder}
\author[2] {Heinz-Peter Schlemmer}
\author[1] {Klaus H. Maier-Hein}
\affil [1] {Division of Medical Image Computing} \affil [2] {Department of Radiology} \affil [3] {Medical Physics in Radiology}
\affil [ ] {German Cancer Research Center, Heidelberg, Germany}

\maketitle

\begin{abstract}
The task of localizing and categorizing objects in medical images often remains formulated as a semantic segmentation problem. This approach, however, only indirectly solves the coarse localization task by predicting pixel-level scores, requiring ad-hoc heuristics when mapping back to object-level scores. State-of-the-art object detectors on the other hand, allow for individual object scoring in an end-to-end fashion, while ironically trading in the ability to exploit the full pixel-wise supervision signal. This can be particularly disadvantageous in the setting of medical image analysis, where data sets are notoriously small.
In this paper, we propose \textit{Retina U-Net}, a simple architecture, which naturally fuses the Retina Net one-stage detector with the U-Net architecture widely used for semantic segmentation in medical images. The proposed architecture recaptures discarded supervision signals by complementing object detection with an auxiliary task in the form of semantic segmentation without introducing the additional complexity of previously proposed two-stage detectors.
We evaluate the importance of full segmentation supervision on two medical data sets, provide an in-depth analysis on a series of toy experiments and show how the corresponding performance gain grows in the limit of small data sets. Retina U-Net yields strong detection performance only reached by its more complex two-staged counterparts. Our framework including all methods implemented for operation on 2D and 3D images is available at \href{https://github.com/pfjaeger/medicaldetectiontoolkit}{github.com/pfjaeger/medicaldetectiontoolkit}.

\end{abstract}

\begin{figure}[t!]
 \includegraphics[scale=0.256]{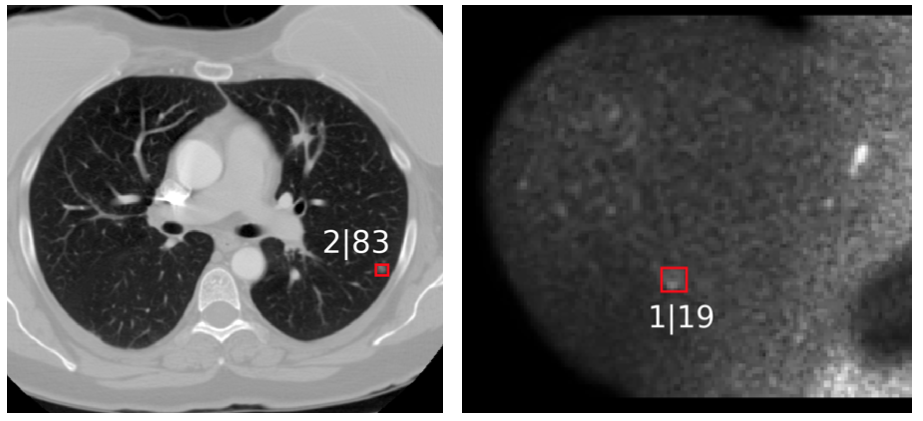}
 \begin{center}
    \caption{Example predictions of a malignant lesion in a CT scan of the lung (left) and a benign lesion on a Diffusion MRI of the breast (right). Object detection in medical images is challenging due to small objects, large images, and limited training data. We tackle this problem by exploiting semantic segmentation supervision in a simple architecture.}
    \label{fig:title}
    \end{center}
\end{figure}

\section{Introduction}
 Semantic segmentation algorithms, such as the U-Net architecture \cite{unet}, constitute the de facto standard approach for detection of anatomical structures in clincial context \cite{medliver, simon, decathlon, medbreast,  medendovis}. This is owed to the circumstance that annotation of medical images is commonly performed by delineating the structures of interest resulting in pixel-wise annotations, which are optimally exploited in the segmentation scenario learning from per pixel supervision signals. Furthermore, MRI and CT imaging capture 3D spaces, inducing inherent spatial separation of objects, therefore not requiring the discrimination of (overlapping) instances. While in tasks like radiation therapy planning or tumor growth monitoring pixel-wise predictions are clinically required, in most settings coarse localization or mere knowledge of object presence are of relevance. This relevance is optimally reflected in study design, when evaluating models on an object-level. In order to bridge the discrepancy between pixel-wise predictions from semantic segmentation and object-level evaluation, however, ad-hoc heuristics or additional models have to be introduced. By extracting predictions from coarser representation levels to enable end-to-end object scoring one naturally converges towards the architecture which most current object detectors are based on: The feature pyramid network (FPN) \cite{fpn}. Two methodologies are applied for generating object-level predictions from coarse representations: two-stage detectors, where first objects are discriminated from background irrespective of class, accompanied by bounding box regression to generate region proposals of variable sizes \cite{maskrcnn, pan, rfcn}. Subsequently, proposals are resampled to a fixed-sized grid to ensure scale-invariance for categorization. On the other hand, one-stage detectors have been proposed, where category-specific classification is performed immediately on the coarse representations \cite{yolo, ssd, retina}. Predicting objects on coarse levels, however, comes at the prize of coarsening pixel-wise annotations to bounding boxes (or cubes). This conversion entails information loss which contradicts the need for data-efficient training on comparably small data sets in the medical domain. We show that fully exploiting the available semantic segmentation signal results in significant performance gains for object detection tasks on medical images. While previous work in the non-medical domain has pursued this concept \cite{priming, pedestrian, megdet, slides}, the extra supervision signal is either used sub-optimally or combined with approaches of arguably significant model complexity. The applicability of computer aided detection systems in clinical environments however, among other requirements, hinges on their interpretability and robustness, which intuitively decreases with model complexity. Furthermore, we argue that the explicit scale variance enforced by the resampling operation in two-stage detectors is not helpful in the medical domain, since unlike in natural images, scale does not depict artifacts caused by varying distances between object and camera, but encodes semantic information.\\\\ Towards the goal of both model simplicity and optimally leveraging available supervision signals, we propose Retina U-Net, a simple approach to recapturing full semantic segmentation supervision based on Retina Net, a plain one-stage detector. Inspired by the U-Net, a very successful model for semantic segmentation of medical images \cite{unet}, we complement the top-down part of the Feature Pyramid Network by additional high resolution levels to learn the auxiliary task of semantic segmentation. From a segmentation perspective, we retrofit the U-Net with two sub-networks operating on the coarse feature levels of the decoder part to allow for end-to-end object scoring.\\\\
We demonstrate the effectiveness of our model on the task of detecting and categorizing lesions on two data sets: Lung-CT (publicly available data \cite{lidc1}) and Breast-Diffusion-MRI (in-house data set described below). We support our analysis by a series of toy experiments that help shed light on the reasons behind the observed performance gains. The proposed model, Retina U-Net, shows results superior to a U-Net like segmentation model with detection heuristics as well as the prevalent object detectors without full semantic supervision: Mask R-CNN \cite{maskrcnn} and Retina Net \cite{retina}. In order to provide a fair and insightful comparison, we enhance the two-stage baselines such that they can be trained with a full segmentation signal and show that Retina U-Net is able to hold up against such more complex models.
This paper makes the following contributions:
\begin{itemize}
    \item A simple but effective method for leveraging semantic segmentation training signals in object detection focused on application in medical images.
    \item An in-depth analysis of the prevalent object detectors (operating in 2D as well as 3D) by means of comparative studies on medical data sets.
    \item Weighted box clustering: An algorithm to consolidate object detections across different predictions of the same image in 2D and 3D.
    \item A comprehensive framework including e.g. modular implementations of all explored models and an efficient implementation of weighted box clustering.
\end{itemize}

\section{Related Work}

\begin{figure*}[!t]
\begin{center}
 \includegraphics[scale=0.24]{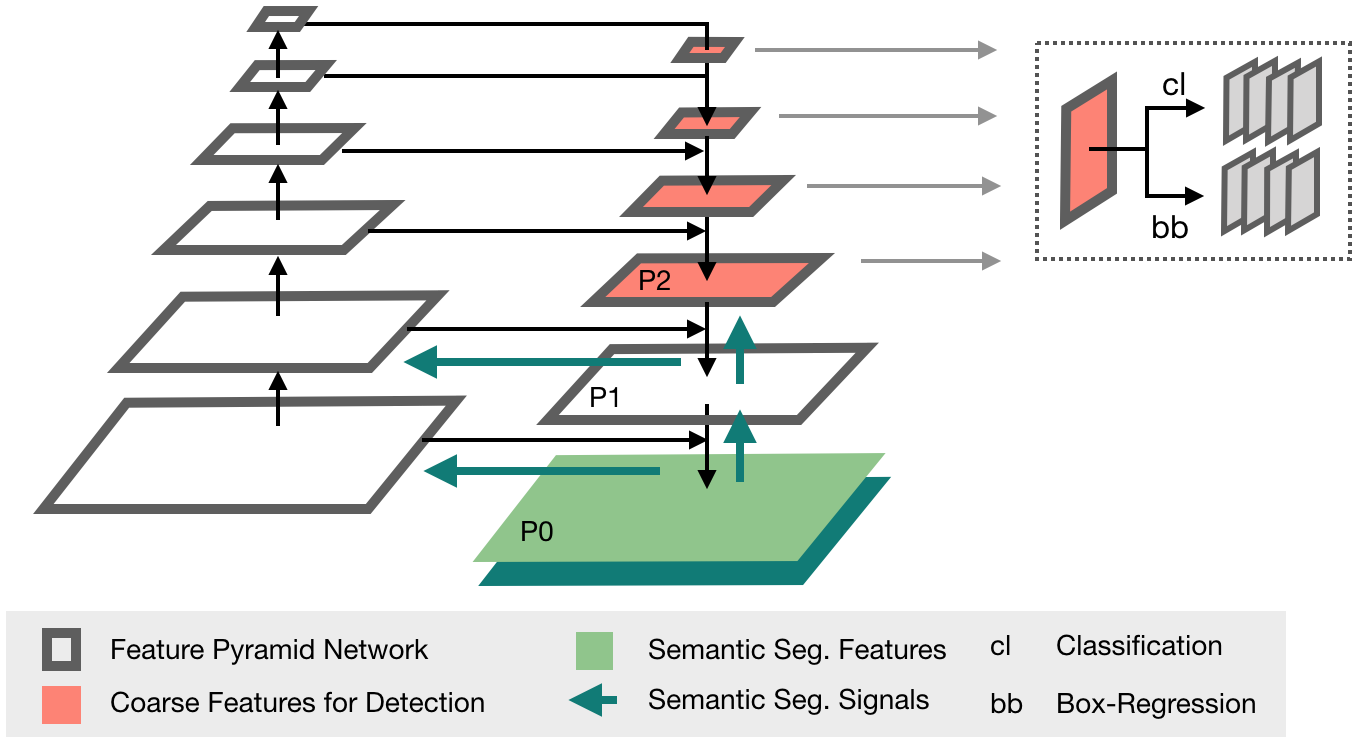}
    \caption{The Retina U-Net architecture in 2D.}
    \label{fig:ret_unet}
\end{center}
\end{figure*}

 Since object detection in natural images is increasingly formulated as an instance segmentation problem, several two-stage object detectors learn to predict proposal-based segmentations \cite{maskrcnn, masklab, pan}. However, we argue that this setup does not fully exploit semantic segmentation supervision:

\begin{itemize}
    \item The mask loss is only evaluated on cropped proposal regions, i.e. context gradients of surrounding regions are not backpropagated.
    \item The proposal region as well as the ground truth mask are typically resampled to a fixed-sized grid (known as RoIAlign \cite{maskrcnn}).
    \item Only positive matched proposals are utilized for the mask loss, which induces a dependency on the region proposition performance.
    \item Gradients of the mask loss do not flow through the entire model, but merely from the corresponding pyramid level upwards.
\end{itemize}
 Auxiliary tasks for exploiting semantic segmentation supervision have been applied in two stage detectors with bottom-up feature extractors (i.e. encoders) \cite{priming, pedestrian}. In the one-stage domain, the work of  Uhrig et al. is the one most similar to ours: Semantic segmentation is performed on top of a single-shot detection (SSD) architecture for instance segmentation, where segmentation outputs are assigned to box proposals in a post-processing step \cite{box2pix}. Zhang et al. propose a similar architecture, but learn segmentation in a weakly-supervised manner, using pseudo-masks created from bounding box annotations \cite{zhang}. \\\\
As opposed to bottom-up backbones for feature extraction, we follow the argumentation of feature pyramid networks \cite{fpn}, where a top-down (i.e. decoder) pathway is installed to allow for semantically rich representations at different scales. This concept is adapted from state-of-the-art segmentation architectures \cite{unet, deeplab} and used in both current one- and two-stage detectors. Building on FPN-based two-stage detectors the winning entries of the COCO object detection challenge 2017 \cite{megdet} and the COCO instance segmentation challenge 2018 \cite{slides} report to have used additional semantic segmentation supervision in increasingly complex backbones like FishNet \cite{slides}, without disclosing implementation details at the time of submission (e.g. the resolution at which segmentation is learned remains unclear). In contrast, we propose a FPN-based one-stage detector, which allows to naturally fuse object detection and segmentation, resulting in the simple Retina U-Net architecture. 

\section{Methods}

\subsection{Retina U-Net}
\label{sec:retu}
\textbf{Retina Net.} Retina Net is a simple one-stage detector based on a FPN for feature extraction \cite{retina}, where two sub-networks operate on the pyramid levels $P_3$-$P_6$ for classification and bounding box regression, respectively. Here $P_j$ denotes the feature-maps of the $j$th decoder level, where $j$ increases as the resolution decreases. To factor in the existence of small object sizes in medical images, we shift sub-network operations by one pyramid level towards $P_2$-$P_5$. This comes at a computational price, since a vast number of dense positions are produced in the higher resolution $P_2$ level. We further exchange the sigmoid non-linearity in the classification sub-network for a softmax operation, to account for mutual exclusiveness of classes due to non-overlapping objects in 3D images. For the 3D implementation, the number of feature maps in the network-heads was reduced to 64, to reduce GPU memory consumption. \\\\
\textbf{Adding Semantic Segmentation Supervision.} In Retina U-Net training signals from full semantic supervision are added to the top-down path by means of additional pyramid levels $P_1$ and $P_0$, including the respective skip connections with the bottom-up path. The resulting Feature Pyramid resembles the symmetric U-Net architecture (see Figure \ref{fig:ret_unet}), which in the following we refer to as \textit{U-FPN} for clarity. The detection sub-networks do not operate on  $P_1$ and $P_0$, which keeps the number of parameters at inference time unchanged. The segmentation loss is calculated from $P_0$ logits. In addition to a pixel-wise cross entropy loss $\mathcal{L_{\mathrm{CE}}}$, a soft Dice loss is applied, which has been shown to stabilize training on highly class imbalanced segmentation tasks e.g. in the medical domain \cite{decathlon}: 
\begin{equation}
\mathcal{L} = \mathcal{L}_\mathrm{CE} - \frac{2}{|K|} \sum_{k\in K}\frac{\sum_{i\in I} u_i^k v_i^k}{\sum_{i\in I} u_i^k + \sum_{i\in I} v_i^k},
\end{equation}
where $u$ is the softmax output of the network and $v$ is a one hot encoding of the ground truth segmentation map. Both $u$ and $v$ have shape $I \times K$ with $i \in I$ being the number of pixels in the training batch and $k\in K$ being the classes. Since dice scores are determined on a per class basis, images with zero foreground pixels in one class lead to unstable scores, where false positive predictions are not penalized.
To alleviate this problem, we compute the dice scores over a pseudo-volume consisting of all images in one batch.

\begin{figure*}

 \begin{center}
 \includegraphics[scale=0.24]{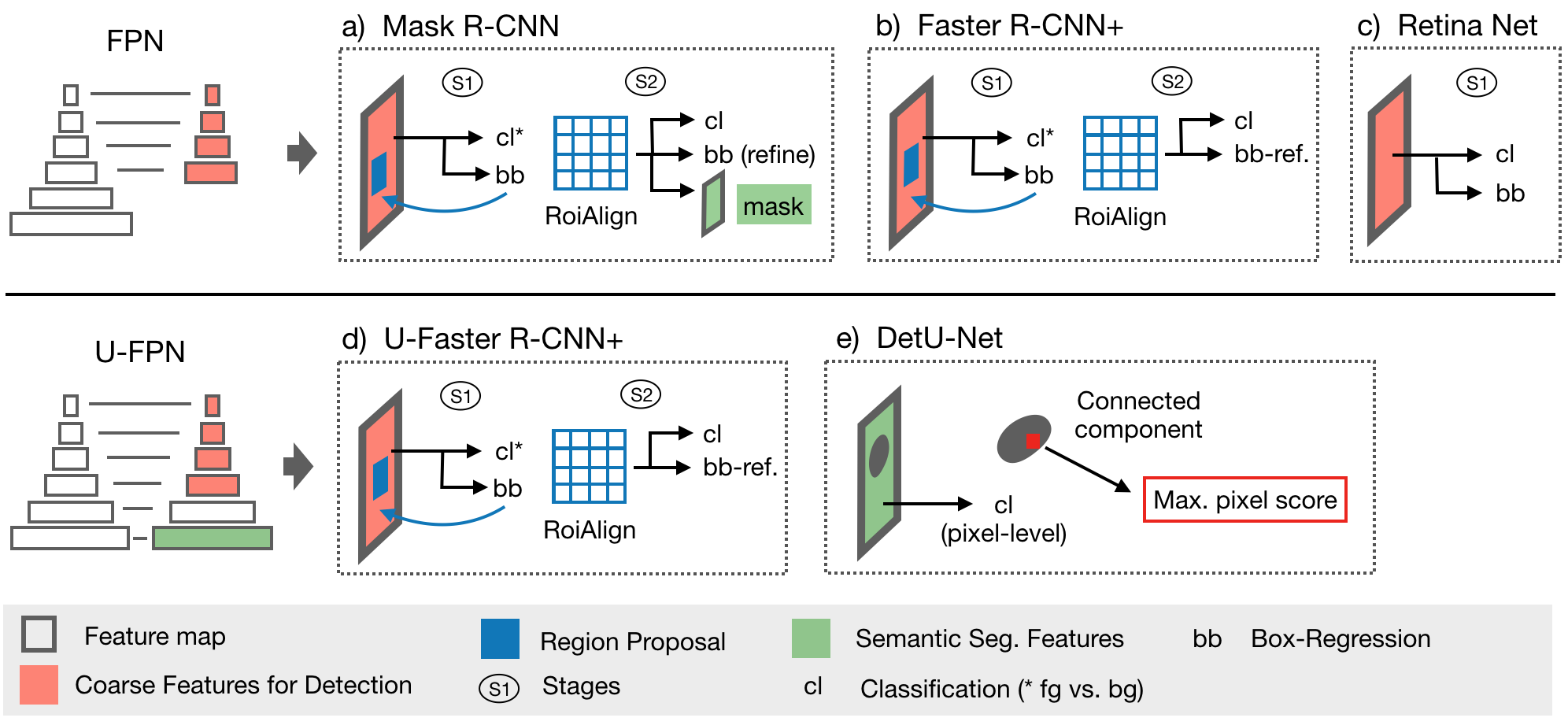}
    \caption{The upper panel shows all baselines utilizing a regular FPN feature extractor while the lower panel depicts baselines that employ a symmetric FPN feature extractor akin to a U-Net (U-FPN). Subfigures a) - e) show the detection sub networks (heads) that are characteristic of each model and operate on FPN features. All models employ their respective head topology to different decoder scales which are denoted in red. Boxes in green indicate logits that are trained on an auxiliary semantic segmentation task.}
    \label{fig:baselines}
\end{center}
\end{figure*}

\subsection{Weighted Box Clustering}
\label{sec:wcs}
\begin{figure}
 \includegraphics[scale=0.515]{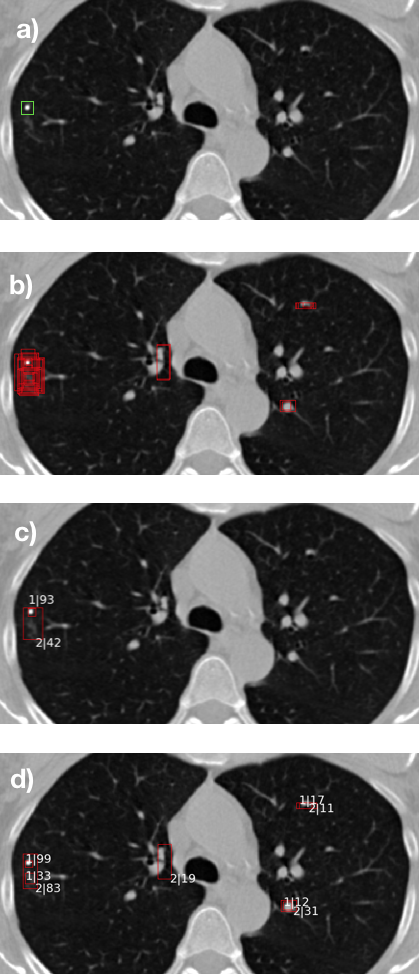}
 \begin{centering}
    \caption{Example application of the \textit{weighted box clustering} (WBC) algorithm. (a) lung CT scan containing a benign lesion marked in green. (b) box predictions of Retina U-Net trained in 2D with context slices: 4 mirror settings from 5 tested epochs lead yield 100 box predictions in the two foreground classes. (c) Remaining predictions after application of WBC: The true lesion is predicted correctly as benign (class 1) with $93\%$ confidence, surrounded by one false positive prediction for class 2 (malignant). (d) Remaining predictions after applying Non-maximum suppression instead of WBC: The benign lesion is predicted correctly, but multiple false positives remain in the image with confidence scores as high as $83\%$.}
    \label{fig:wcs}
    \end{centering}
\end{figure}

Medical images typically are comparatively large, due to e.g. very high resolutions as in mammograms, or since they are acquired as 3D volumes like in MRI. Image resolutions are expected to keep rising in the future driven by advances in imaging technologies, such as the recent introduction of 7T MRI scanners. For this reason models are trained on patch crops, resulting in a trade-off between patch size and batch size limited by available GPU memory. If a single image exceeds GPU memory, inference is performed patch-wise as well, where tiling strategies are designed to avoid potential artifacts that arise due to effects at the patch boundaries (e.g. by allowing for sufficient overlap between patches). The tiling strategies as well as test time augmentations and model ensembling can amount to a large number of predictions per patch and image (particularly in medical object detection, where validation metrics for model selection are often based on limited validation data, ensembling over multiple selected epochs is able to reduce noise effects in this process). The resulting predictions for different views of the same image need to be consolidated, which in semantic segmentation is done via simple per-pixel averaging. For consolidation of predictions in object detection, we propose \textit{weighted box clustering} (WBS):  Similar to the commonly used non-maximum suppression algorithm, WBS clusters predictions to be consolidated according to an IoU threshold, but instead of selecting the highest scoring box in the cluster, weighted averages $o_c$ per coordinate and a weighted confidence score $o_s$ per resulting box are computed. Further, the prior knowledge about the expected number of predictions at a position (number of views from ensembling, test time augmentations and patch overlaps at the position) is used to down-weight $o_s$ for views that did not contribute at least one box to the cluster ($n_{missing}$):
\begin{equation}
    o_s = \frac{\sum s_i w_i}{\sum w_i + n_{missing} * \bar{w}}, \quad o_c = \frac{\sum c_i s_i w_i}{\sum s_i w_i}, 
\end{equation}
where $i$ is the cluster members' index, $s$ and $c$ the corresponding confidence scores and coordinates. $w = f \cdot a \cdot p $ is the weighting factor, consisting of:
\begin{itemize}
    \item overlap factor $f$: weights according to the overlap between a box and the highest scoring box (softmax confidence) in the cluster.
    \item area $a$: assigns higher weights to larger boxes based on empirical observations indicating an increase in image evidence from larger areas. 
    \item patch center factor $p$: down-weights boxes based on the distance to the input patch center, where most image context is captured. Scores are assigned according to the density of a normal distribution that is centered at the patch center.
\end{itemize}
An example for application of WBC is shown in Figure \ref{fig:wcs}.

\section{Experiments}

\subsection{Backbone \& Baselines}
 In this study, we compare Retina U-Net against a set of one- and two-stage object detectors. We evaluate the performance of all models processing the data in both 2D (slice-based) and 3D (volumetric patches). For the sake of unrestricted comparability, all methods are implemented in one framework, using a FPN \cite{fpn} based on a ResNet50 backbone \cite{resnet50} as identical architecture for feature extraction. The anchor sizes are divided by a factor of 4 to account for smaller objects in the medical domain resulting in anchors of size \{$4^2, 8^2, 16^2, 32^2$\} for the corresponding pyramid levels \{$P_2, P_3, P_4 , P_5$\}. In the 3D implementation, the z-scale of anchor-cubes is set to \{$1, 2, 4, 8$\} factoring in the typically lower resolutions along the z-axis.\\\\
\textbf{Retina Net}. The implementation of Retina Net is identical to the one used in Retina U-Net and described in \ref{sec:retu} (see Figure \ref{fig:baselines}c).\\\\
\textbf{Mask R-CNN}. Minor adjustments have to be made for the 3D implementation: The number of feature maps in the region proposal network is lowered to 64 to reduce GPU memory consumption. The poolsize of 3D-RoIAlign is set to (7, 7, 3) for the classification head and (14, 14, 5) for the mask head. The matching IoU for positive proposals is lowered to 0.3 (see Figure \ref{fig:baselines}a). \\\\ 
\textbf{Faster R-CNN+}. In order to single out the Mask R-CNN's performance gain obtained by segmentation supervision from the mask head, we run ablations on the toy data sets while disabling the mask-loss, thereby effectively reducing the model to the Faster R-CNN architecture  \cite{frcnn} except for the RoIAlign operation (which we denote by a +) (see Figure \ref{fig:baselines}b).\\\\
\textbf{U-Faster R-CNN+}. As the currently leading COCO challenge architecture's exact implementation details are presently undisclosed \cite{slides}, we explore the performance of additional semantic segmentation in two-stage detectors by deploying Faster R-CNN+ on top of U-FPN. (see Figure \ref{fig:baselines}d).\\\\
\textbf{DetU-Net}. Essentially formulating the problem as a semantic segmentation task, as is common on medical imaging, we implement a U-Net-like baseline using U-FPN. Therefore softmax predictions are extracted from $P_0$ via 1x1 convolution and utilized to identify connected components for all foregorund classes. Subsequently, bounding boxes (or cubes) are drawn around connected components and the highest softmax probability per component and class is assigned as object score. To reduce noise, only the 5 (15 in 3D) largest components per image are considered (see Figure \ref{fig:baselines}e).\\\\

\begin{table}[!t]
  \caption{Results for lung lesion detection on CT.}
  \label{tab:lidc}

  \centering
  \begin{tabular}{llll}
    \\
    Dim. & Model     & $mAP_{10}$  [\%]    & $AP_{pat_m}$ [\%] \\
    \toprule
    & Retina U-Net & 49.8 & 70.4      \\
   & Retina Net & 45.9 & 68.8      \\
    3D & DetU-Net & 36.6 & 62.8      \\
    & U-FRCNN+ & \textbf{50.5} & 70.7      \\
    & Mask R-CNN & 48.3 & \textbf{71.8}      \\
    \midrule
     & Retina U-Net & \textbf{50.2}   & \textbf{73.9}    \\
    & Retina Net & 48.2 & 71.5    \\
   2Dc      & DetU-Net & 41.1 & 66.1      \\
    & U-FRCNN+ & 49.1   & 71.6  \\
    & Mask R-CNN & 45.4 & 69.1     \\
    \midrule
    & Retina U-Net & \textbf{40.8} & \textbf{68.0}      \\
   & Retina Net & 39.5 & 67.7      \\
   2D  & DetU-Net & 29.0   &    59.8 \\
    & U-FRCNN+ & 38.6 & 66.4      \\
    & Mask R-CNN & 35.3 & 63.6      \\

    \bottomrule
  \end{tabular}

  \end{table}
  
\begin{table}[!t]

\caption{Results for breast lesion detection on Diffusion-MRI.}
\label{tab:breast}
  \centering
   \begin{tabular}{llll}
    \\
    Dim. & Model     & $mAP_{10}$ [\%]     & $AP_{pat_m}$ [\%] \\
    \toprule
     & Retina U-Net & \textbf{35.8}   & \textbf{88.0}    \\
   &  Retina Net & 31.9 & 86.4     \\
     3D    & DetU-Net & 26.9 & 85.1      \\
    & U-FRCNN+ & 35.1 & 86.5      \\
    &  Mask R-CNN & 34.0 & 84.8   \\
    
    \midrule
    & Retina U-Net & \textbf{33.4} & \textbf{86.9}      \\
    & Retina Net & 33.2 & 84.4     \\
  2Dc     & DetU-Net & 25.8  & 81.6\\
    & U-FRCNN+ & 33.2 & 84.7      \\
    & Mask R-CNN & 32.3 & 86.4      \\
   
    \midrule
    &      Retina U-Net & 31.8 & 85.2  \\
    & Retina Net & 32.3 & 86.1      \\
 2D      & DetU-Net & 22.3 & 80.1      \\
    & U-FRCNN+ & 33.1 & 83.7      \\
    &     Mask R-CNN & \textbf{33.6} & \textbf{86.8}  \\
    \bottomrule
  \end{tabular}

  \label{tab:results}
\end{table}

\subsection{Training \& Inference Setup}  For comparability, experiments for all methods are run with identical training and inference schemes. Since experiments are performed on 3D images, several possible ways of processing the volumetric data arise. In this study, we compare slice-wise 2D processing, slice-wise 2D processing feeding the $\pm 3$ neighbouring slices as additional input channels (2Dc) \cite{ron}, and 3D volume processing (i.e. using volumetric convolutions). Oversampling of foreground regions is applied when training on patch crops. To account for the class-imbalance of object level classification losses, we stochastically mine the hardest negative object candidates according to softmax probability. Models are trained in a 5-fold cross validation (splits: train 60\% / val 20\% / test 20\%) with batch size 20 (8) in 2D (3D) using the Adam \cite{adam} optimizer at a learning rate of $10^{-4}$. Extensive data augmentation in 2D and 3D is applied to account for overfitting. To compensate for unstable statistics on small data sets, we report results on the aggregated inner loop test sets and ensemble by performing test-time mirroring as well as by testing on multiple models selected as the 5 highest scoring epochs according to validation metrics. Consolidation of box predictions from ensemble-members and overlapping tiles is done via clustering and weighted averaging of scores and coordinates, as detailed in Section \ref{sec:wcs}. Since evaluation is performed entirely in 3D, a adaption of non-maximum suppression (NMS) is applied to consolidate box predictions from 2D networks to 3D cube predictions: Boxes of all slices are projected into one plane while retaining the slice-origin information. When applying NMS, only boxes with direct or indirect connection to the slice of the highest scoring box are considered as matches. The minimal and maximal slice numbers of all matches are assigned as z-coordinates to the resulting prediction cube. 

\subsection{Evaluation} Experiments are evaluated using mean average precision (mAP). We determine mAP at a relatively low matching intersection over union (IoU) threshold of $\SI{}{IoU} = 0.1$. This choice respects the clinical need for coarse localization and also exploits the non-overlapping nature of objects in 3D. Note, that evaluation and matching is performed in 3D for all models and processing setups. Patient-level scoring is often used in the clinical context for staging and treatment and is therefore sometimes used for model selection. In our setting the patient-level metric is to be taken with a grain of salt: By disregarding whether box predictions match the ground truth, it is blind to the issue of `being right for the wrong reasons' and can further over-estimate performance due to class-imbalance on the patient-level. For the purpose of comparability, we however also report patient-scores in Table \ref{tab:results} which are determined as the maximum of predicted scores per class and patient and compute the AP thereof.

\subsection{Lung nodule detection and categorization}
We consider the task of detecting lesions and assigning them to one of two categories, \textit{benign} or \textit{malignant}. This is both a difficult and very frequent problem setting in radiology and therefore constitutes a highly-relevant domain of application with its own characteristics that our approach factors in and addresses. Furthermore, fine-grained categorization is expected to gain relevance in the context of growing data sets and image resolutions. 
\subsubsection{Utilized data set}
The lung nodule detection and categorization task is performed on the publicly available LIDC-IDRI data set \cite{lidc1, lidc2, lidc3}, consisting of 1035 lung CT scans with manual lesion segmentations and malignancy likelihood scores (1-5) from four experts. We aggregate labels across raters per lesion by applying a pixel-wise majority voting on the segmentations and taking the mean of malignancy scores. Scores are then re-labelled into benign (1-2, n=1319) and malignant (3-5, n=494). CT scans are resampled to a resolution of $0.7 \times 0.7 \times \SI{1.25}{\milli\meter}$, which roughly corresponds to the mean resolution of the data set. For training, patches of size $288 \times 288$ (for 3D training: $128 \times 128 \times 64$) are sampled.

\begin{figure*}[!t]

 \includegraphics[scale=0.29]{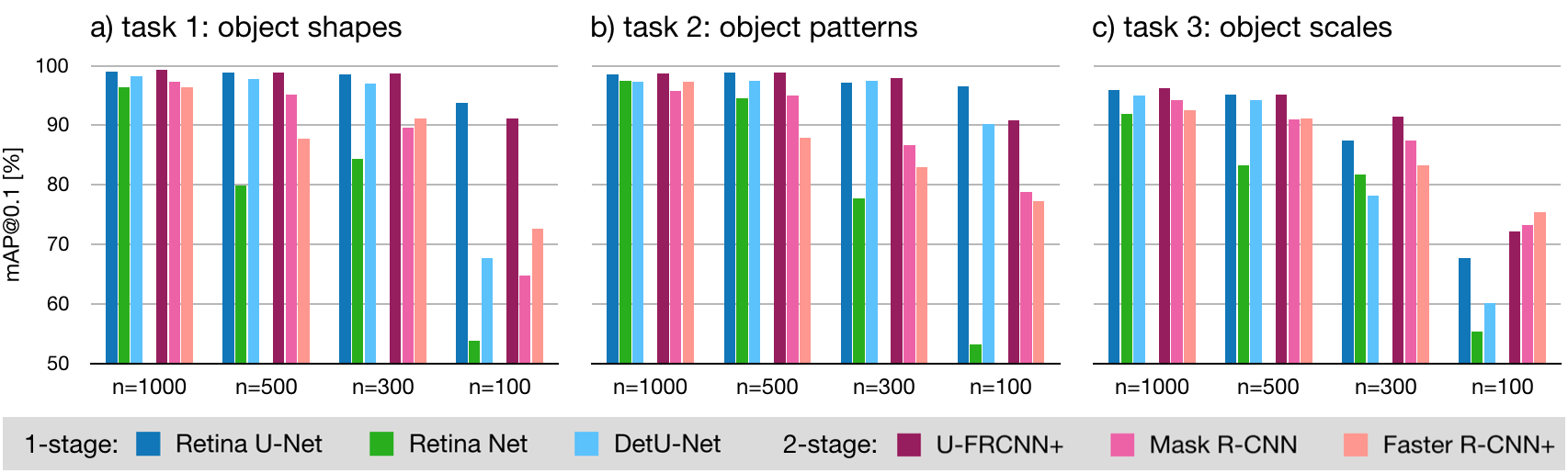}
 \begin{centering}
    \caption{Results of the of toy experiment series. The three tasks are displayed as (a) distinguishing objects of different shapes, (b) learning discriminative image patterns unrelated to an object's shape, and (c) distinguishing objects of different scales. Explored models are divided into two groups: One-stage methods have blue/green color, while two-stage methods are drawn in red. $n$ denotes the number of utilized training images.}
    \label{fig:toy_results}
    \end{centering}
\end{figure*}

\subsubsection{Results}
Results for the lung lesion detection task are shown in Table \ref{tab:lidc}. Retina U-Net performs best on the 2D setups (\SI{0.41}{mAP} and \SI{50.2}{mAP} with context slices) and only slightly worse (\SI{0.50}{mAP}) than its two-stage counterpart (\SI{0.51}{mAP}), the U-FRCNN+ in 3D. Comparing the two to the remaining baselines shows a clear performance margin, hinting upon the importance of full segmentation supervision. The DetU-Net performs worse with a notable margin (\SI{0.41}{mAP} in 2D with context slices and \SI{0.37}{mAP} in 3D), seemingly suffering from high confidence false positive predictions caused by the ad-hoc score aggregation. Generally, 3D context shows to be important for this task, yet operating entirely in 3D seems to yield no benefits with respect to feeding the $\pm$ 3 neighbouring slices to a 2D network.

\subsection{Breast lesion detection and categorization}
\subsubsection{Utilized data set}
 The breast lesion detection and categorization task is performed on an in-house Diffusion MRI data set of 331 Patients with suspicious findings in previous mammography. Lesion annotations are provided by experts. Categorization labels are given by subsequent pathology results (benign: n=141, malignant: n=190). Images are resampled to a resolution of $1.25 \times 1.25 \times \SI{3}{\milli\meter}$. For training, images are cropped to a size of $160 \times 160$ (for 3D training:  $160 \times 160 \times 56$).
 
\subsubsection{Results}
Results for the breast lesion detection task are shown in Table \ref{tab:breast}.
 Retina U-Net performs best in 2D with context slices (\SI{0.33}{mAP}) and in 3D (\SI{0.36}{mAP}). Mask R-CNN yields best results on plain 2D (\SI{0.34}{mAP}). The overall best results are achieved by Retina U-Net and U-FRCNN+, again indicating that leveraging the full segmentation information is crucial in medical object detection. As opposed to the lung CT data set, 3D context information seems to be less important here, which is expected, when operating on highly anisotropic data, i.e. data with low z-resolution. Notably, the patient-level scores are only poorly correlated to the actual model performance due to high ambiguities in score aggregation and should not be used for model selection in clinical context.

\subsection{Toy Experiments}
\begin{figure}

 \includegraphics[scale=0.29]{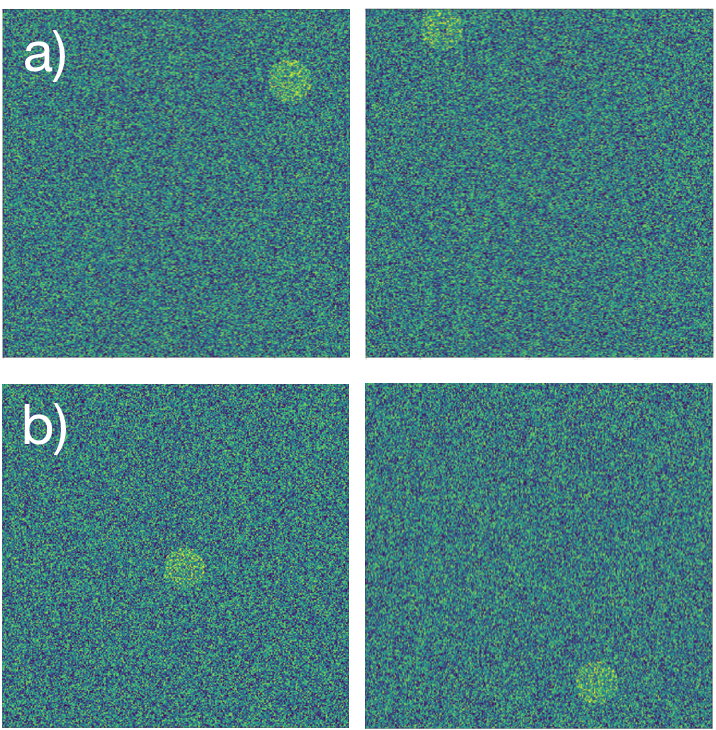}
 \begin{centering}
    \caption{Image samples of the toy experiments. (a) the first two tasks require to detect and distinguish filled circles (left) from donuts (right) over noisy background. (b) For the third task, filled circles of different scales have to be distinguished: 19 pixels diameter (left) versus 20 pixels diameter (right).}
    \label{fig:toy_examples}
    \end{centering}
\end{figure}

We create a series of toy experiments to get a handle on the sub-tasks commonly involved in object-categorization on medical images, such as distinguishing scales, shapes and intensities. More specifically, the aim is to investigate the importance of full segmentation supervision in the context of limited training data. Therefore we consider three tasks for each of which we gradually decrease the amount of training data:
\begin{enumerate}
    \item \textbf{Distinguishing object shapes:} Two classes of objects are to be detected and distinguished, where class 1 consists of filled circles with 20 pixels diameter and class 2 consists of filled circles with 20 pixels diameter and a centered whole of 4 pixels diameter, resembling the shape of donuts (see Figure \ref{fig:toy_examples}a). For this task, the corresponding segmentation mask's shape explicitly contains the discriminative feature, since the centered whole is cut out from the foreground mask. Hence, full semantic supervision is expected to yield significant performance gains on this task, particularly in the small data regime.
    \item \textbf{Learning discriminative object patterns:} This task is identical to the previous one, except the central hole is not cut out from the segmentation masks of the donuts (class 2). This requires the model to pick up the discriminative pattern (the hole) without explicitly pointing it out by means of the mask's shape (see Figure \ref{fig:toy_examples}a). This setup could be considered more realistic in the context of medical images.
     \item \textbf{Distinguishing object scales:} Two classes of objects are to be detected and distinguished, where class 1 consists of circles with 19 pixels diameter and class 2 consists of circles with 20 pixels diameter (see Figure \ref{fig:toy_examples}b). Here, class information is entirely encoded in object scales and hence in target box coordinates. No significant gain from semantic supervision is expected.
\end{enumerate}

\subsubsection{Utilized data set}
Both data sets consist of artificially generated 2D images of size $320 \times 320$, where 1000 images were created for training, 500 for validation, and another 1000 as held-out test set. Images are zero-initialized and foreground objects imprinted by increasing intensity values by 0.2. Subsequently, uniform noise is added to all pixels.  
\subsubsection{Results}
Results are shown in Figure \ref{fig:toy_results}. In the first task, where explicit class information is contained in segmentation annotations, models which explicitly leverage those, i.e. Retina U-Net and U-FRCNN, perform best. The resulting margin increases with decreasing amount of available training data. The second task, where class information is effectively removed from segmentation annotations, similar margins of Retina U-Net and U-FRCNN+ towards the other models are observed. This indicates the importance of full segmentation supervision even in implicit setups and shows a particularly strong edge in the small data set regime, where models that discard this supervision essentially collapse. In the third task, where class information is entirely contained in the target boxes, no gain from segmentation supervision is observed, at least for small training data sets. Interestingly, two-shot detectors perform better at this task, which seems counter-intuitive given the scale-invariance enforced by the RoIAlign operation. We hypothesize, that this bottleneck regularizes the architecture in a sense that scale information is enforced to be encoded broadly over spatial dimensions of FPN features.
Comparing Mask R-CNN to Faster R-CNN+, the sub-optimal mask-supervision seems to yield no gains in detection performance when working with limited training data.

\section{Conclusion}
In this work, we propose Retina U-Net, a simple but effective one-stage detection model which leverages semantic segmentation inspired by the current state-of-the-art architecture in medical segmentation. We show the importance of exploiting these training signals on multiple data sets, input dimensions and meticulously compare against the prevalent object detection models, with a particular emphasis on the context of limited training data. Therefore, we consider the task of localizing and classifying lesions, which constitutes a difficult and very frequent problem setting in radiology and therefore a highly-relevant domain of application. On the publicly available LIDC-IDRI lung CT dataset as well as on our in-house breast lesion MRI dataset Retina U-Net yields detection performance superior to models without full segmentation supervision and only reached by its more complex two-stage counterpart. By means of a set of toy experiments we shed light on a important set of scenarios that can profit from the additional full supervision: Any such problem where there is discriminative power in features beyond mere scale can expect to pocket an edge in detection performance. Among other distinguishing characteristics, the domain of medical image analysis holds one prominent feature: scarcity of labelled data. Retina U-Net is designed to make the most of the given supervision signal which is a key advantage on small datasets as high-lighted by our experiments. Our architecture stands out with another feature: its embarrassingly simple model formulation that takes aim at the clinical requirement of interpretability and robustness.




\end{document}